\documentclass{bmvc2k}

\usepackage{booktabs}
\usepackage{pifont}
\usepackage{booktabs}
\usepackage{amsfonts}
\usepackage{multicol}
\usepackage{multirow}
\usepackage{array}
\usepackage{amsmath}
\usepackage{appendix}
\usepackage[hang,flushmargin]{footmisc}

\title{Zero-shot Composed Text-Image Retrieval}

\addauthor{Yikun Liu}{yikunliu@sjtu.edu.cn}{1,2}
\addauthor{Jiangchao Yao}{Sunarker@sjtu.edu.cn}{1,3}
\addauthor{Ya Zhang}{ya_zhang@sjtu.edu.cn}{1,3}
\addauthor{Yanfeng Wang}{wangyanfeng622@sjtu.edu.cn}{1,3, $^\dagger$}
\addauthor{Weidi Xie}{weidi@sjtu.edu.cn}{1,3, $^\dagger$}

\addinstitution{
 Coop. Medianet Innovation Center,\\
 Shanghai Jiao Tong University, China
}
\addinstitution{
 Beijing University of Posts and\\
 Telecommunications, China
}
\addinstitution{
 Shanghai AI Laboratory
}

\runninghead{LIU, ET.AL.}{ZERO-SHOT COMPOSED TEXT-IMAGE RETRIEVAL}

\begin{document}

\maketitle

\begin{abstract}
In this paper, we consider the problem of composed image retrieval~(CIR), it aims to train a model that can fuse multi-modal information, 
{\em e.g.}, text and images, to accurately retrieve images that match the query, extending the searching ability. 
We make the following contributions: 
(i) we initiate a scalable pipeline to automatically construct datasets for training CIR model, by simply exploiting a large-scale dataset of image-text pairs, {\em e.g.}, a subset of LAION-5B;
(ii) we introduce a transformer-based adaptive aggregation model, 
\textbf{TransAgg}, which employs a simple yet efficient fusion mechanism, 
to adaptively combine information from diverse modalities; 
(iii) we conduct extensive ablation studies to investigate the usefulness of our proposed data construction procedure, and the effectiveness of core components in TransAgg; 
(iv) when evaluating on the publicly available benchmarks under the zero-shot scenario, {\em i.e.}, training on the automatically constructed datasets, 
then directly conduct inference on target downstream datasets, 
{\em e.g.}, CIRR and FashionIQ, our proposed approach either performs on par with or significantly outperforms the existing state-of-the-art (SOTA) models. Project page: \href{https://code-kunkun.github.io/ZS-CIR/}{https://code-kunkun.github.io/ZS-CIR/}

\end{abstract}

\section{Introduction}
\label{sec:intro}

In the recent literature, vision-language models have made tremendous progress, by jointly training image and text representation on large-scale dataset collected from the Internet. For example, CLIP~\cite{radford2021learning} and ALIGN~\cite{jia2021scaling} trained with simple noise contrastive estimation~\cite{oord2018representation}, 
have demonstrated surprisingly strong transferability and generalizability 
on zero-shot classification or cross-modal retrieval. 
In this paper, we consider the task of composed image retrieval~(CIR), 
that aims to retrieve images by leveraging a combination of reference image and textual information that illustrates desired modifications. 
The model needs to use visual and language representation interchangeably, 
and discover target images that satisfy the user's expectation. 
In comparison to image-to-image or text-to-image retrieval, 
CIR captures richer semantics about the user's intention,
and thus has the potential to enable more precise retrieval on images or e-commerce products.

Existing approaches~\cite{baldrati2022conditioned,liu2021image,delmas2022artemis,vo2019composing} for composed image retrieval typically train deep neural networks under fully supervised setting, which requires a dataset, consisting of sufficient \{a reference image, a relative caption, and a target image\} triplets. However, compared with collecting the text-image pairs, manually constructing such a triplet dataset is usually very expensive, that requires substantial human efforts, to thoroughly examine the reference image and target image and produce a text description to capturie their distinctions. 
Consequently, the practical datasets for training CIR models tend to be limited by scale. 

In this paper, 
we initiate a scalable pipeline to automatically construct datasets for training CIR model, 
by exploiting the vast amount of image-caption data available on the Internet. Specifically, for one image-caption sample, we can revise its caption and use the resulting edited caption as a query to retrieve the target image with similar caption, 
where we adopt an off-the-shelf Sentence Transformer to compute similarity between sentences. Depending on the different approaches for revising captions,
{\em i.e.}, using template or large language models~(LLM), we obtain two different training datasets respectively. In addition, we introduce a transformer-based model, that employs a simple yet efficient fusion mechanism to adaptively combine information from diverse modalities. Once trained on the automatically constructed datasets, 
the model can be directly applied to target downstream CIR benchmarks without any finetuning, thus advocates zero-shot generalisation.

To summarise, we make the following contribution: 
(i) we propose a retrieval-based pipeline for automatically constructing dataset for training, with the easily-acquired image-caption data on Internet; 
(ii) we introduce a transformer-based aggregation model, 
termed as \textbf{TransAgg}, that employs a simple yet efficient modules to dynamically fuse information from different modalities. (iii) we train a model on the automatically constructed dataset, and directly evaluate on publicly available CIR benchmarks,
thus resembling zero-shot composed image retrieval. In particular, we extensively evaluate the applicability of our constructed dataset, with different pre-trained backbones and fine-tuning types, and perform thorough ablation studies to validate the effectiveness of the transformer module and adaptive aggregation of our model; (iv) while comparing with existing approaches on two public benchmarks under zero-shot scenario, namely, CIRR and FashionIQ, our model performs on par or significant above the existing state-of-the-art (SOTA) models, and is sometimes comparable to fully supervised ones.


\section{Related Work}
\label{sec:related_work}


\noindent \textbf{Image Retrieval. } 
Standard image retrieval includes both image-to-image retrieval and text-to-image retrieval. Existing research can be mainly divided into two categories. 
One uses dual tower structure~\cite{pan2016jointly,miech2019howto100m,dong2019dual,klein2015associating}. 
It relies on a good feature extractor to get features of text or image, and then uses cosine similarity for retrieval. 
The other one is to pass image-image or text-image pairs through a mutli-modal encoder to compute their similarity~\cite{ni2021m3p,li2020oscar,bugliarello2021multimodal}. 
Despite the impressive progress, 
these retrieval models are unable to exploit the complemantary information in different modalities for constructing fine-grained queries.


\vspace{3pt}
\noindent \textbf{Composed Image Retrieval. } Composed Image Retrieval (CIR) considers the problem of retrieving images based on the reference images and relative captions. 
Till recently, majority research in CIR has concentrated on the fusion of multiple modalities to generate optimal multimodal representations.
Specifically, TIRG~\cite{vo2019composing} proposes to use residual modules and gating modules to fuse features. CIRPLANT~\cite{liu2021image} employed vision-and-language pre-trained (VLP) multi-layer transformers to fuse features that come from distinct modalities. CLIP4CIR~\cite{baldrati2022conditioned} leverages CLIP~\cite{radford2021learning} as feature extractor and follows a two-stage training procedure. In the first stage, CLIP~\cite{radford2021learning} text encoder is fine-tuned, and a combiner is trained in the second stage, culminating in remarkable outcomes.

\vspace{3pt}
\noindent \textbf{Concurrent Work. } 
Several recent papers~\cite{saito2023pic2word,gu2023compodiff,levy2023data} also explore the idea of zero-shot composed image retrieval, specifically, Pic2Word~\cite{saito2023pic2word} employs image-caption and unlabeled image datasets to train a mapping network that marks the image as a token, and performs cross-modal retrieval with CLIP~\cite{radford2021learning}.  
CompoDiff~\cite{gu2023compodiff} proposes a two-stage approach for training diffusion model to address the CIR problem and introduces the SynthTriplet18M dataset, comprising images synthesized via the prompt-to-prompt~\cite{hertz2022prompt} model guided by corresponding captions.  CASE~\cite{levy2023data} proposes to use BLIP~\cite{li2022blip} model to accomplish the CIR task through early fusion and utilzing the few-shot capability of GPT-3\cite{brown2020language}, and the VQA2.0\cite{goyal2017making} dataset to construct a dataset of almost 400K triplets in an semi-automatic manner. The target images are manually selected from the 24 visually nearest neighbors of referenece images. Unlike the aforementioned approach, our approach is fully automated and does not require any human intervention, based on retrieval from a large-scale corpus of real images.

\vspace{-5pt}
\section{Method}

In this section, we start by formulating the problem of composed image retrieval in Sec.~\ref{sec:problem}, 
then provide details of our proposed architecture in Sec.~\ref{sec:model}, 
lastly, in Sec.~\ref{sec:dataset_construction}, 
we describe the two ideas for automatically constructing training set for CIR task, 
namely, Laion-CIR-Template and Laion-CIR-LLM.

\vspace{-0.1cm}
\subsection{Problem Scenario}
\label{sec:problem}

We consider the problem of composed image retrieval, 
specifically, at training time, each sample can be represented as a triplet, 
{\em i.e.}, $\mathcal{D}_{\text{train}} =\left\{\left( I_{r},I_{t}, t \right)| I_{r} \in \mathbb{R}^{H \times W \times 3}, I_{t} \in \mathbb{R}^{H \times W \times 3}\right\}$, 
Specifically, we train a model that takes the reference image~($I_r$) and 
relative caption~($t$) as input, and construct a composed query, that can retrieve one target image~($I_t$):
\begin{equation}
\small
    Q = \Phi_{\text{TransAgg}}(I_r, t) = \Phi_{\text{agg}}(\Phi_{\text{fuse}}(\hspace{1pt} \Phi_{\text{visual}}\left( I_{r} \right), 
               \hspace{3pt} \Phi_{\text{text}}\left( t \right) \hspace{1pt}))
\end{equation}
$Q$ refers to the composed query, 
that is to rank all images in a retrieval set $\Omega = \{I_i, i = 0, \cdots , m\}$ based on the relevance, {\em i.e.}, cosine similarity computed by between query and image embedding. For each composed query, the retrieval set is split into positive $P_q$ and negative $N_q$ sets, with the former consisting of instances that satisfy conditional editing on reference image. The trainable modules include: visual encoder~($\Phi_{\text{visual}}$), text encoder~($\Phi_{\text{text}}$), multi-modal fusion module~($\Phi_{\text{fuse}}$), and an aggregation module~($\Phi_{\text{agg}}$). 

\vspace{-0.1cm}
\subsection{Composed Image Retrieval Model}
\label{sec:model}

\begin{figure}
    \centering
    \includegraphics[width=.97\textwidth]{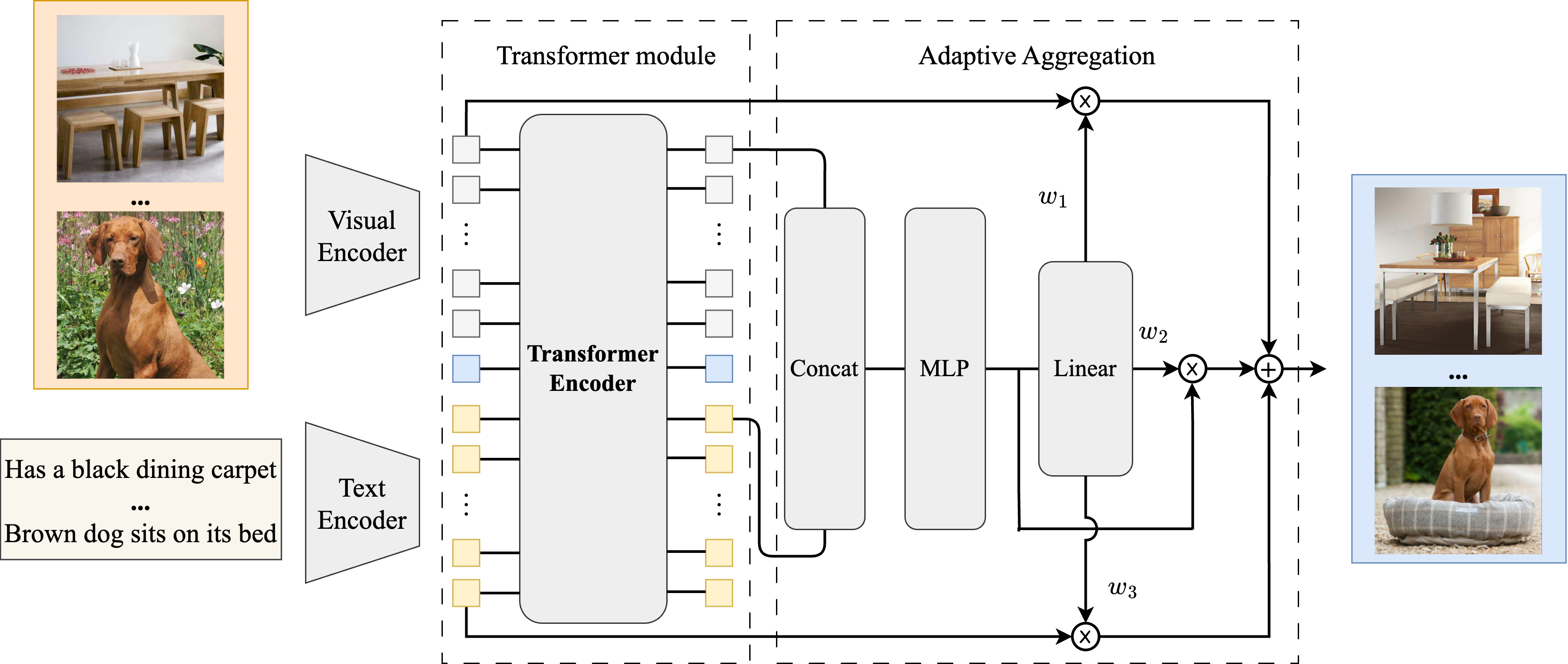}
    \caption{An overview of our proposed architecture, that consists of a visual encoder, 
    a text encoder, a Transformer module and an adaptive aggregation module. }
    \label{fig:architecture}
\end{figure}


Here, we start by introducing our proposed model for composed image retrieval, termed as TransAgg, 
and followed by its detailed training objective.

\subsubsection{Architecture} \label{sec:arch}

As illustrated in Figure \ref{fig:architecture}, 
our proposed CIR model consists of three components: 
encoders to extract features from visual and textual inputs respectively, 
a Transformer module to capture the interaction between two modalities, 
and an adaptive aggregation module that combats modal redundancy and fuses the features together. 

\vspace{3pt}
\noindent \textbf{Visual and Text Encoders.} 
We adopt pre-trained vision and language models as our encoders for different modalities given their impressive performance and flexibility to maintain the semantics. Formally, we denote the feature extraction via the following notations,
\begin{equation}
    \small
    \begin{split}
        & \mathcal{F}_{\mathrm{Vr}} =\Phi_{\text{visual}}\left( I_{r} \right)\in \mathbb{R}^{|\mathcal{V}| \times d}, \qquad \mathcal{F}_{\mathrm{W}} =\Phi_{\text{text}}\left( t \right)\in \mathbb{R}^{|\mathcal{W}| \times d}
    \end{split}
\end{equation}
where $I_{r}$ denotes the reference image encoded by the visual encoder $\Phi_{\mathrm{visual}}$, and $t$ refers to the relative caption encoded by the textual encoder $\Phi_{\text{text}}$. 
In our experiments, we primarily use pretrained BLIP~\cite{li2022blip} or CLIP~\cite{radford2021learning} as our visual and text encoders.

\vspace{5pt}
\noindent \textbf{Transformer Fusion.}
Regarding the input of our Transformer module, 
in addition to $\mathcal{F}_{\mathrm{Vr}}$ and $\mathcal{F}_{\mathrm{W}}$, 
a learnable token embedding $\mathcal{F}_{\mathrm{sep}} $ is also integrated to discriminate the modalities. The feature interaction between visual and textual modality can be formulated as:
\begin{equation}
\small
\left[\mathcal{F}_{\mathrm{Vr}}^{\prime}, \mathcal{F}_{\mathrm{sep}}^{\prime}, \mathcal{F}_{\mathrm{W}}^{\prime}\right]=\Phi_{\mathrm{fuse}}\left(\left[\mathcal{F}_{\mathrm{Vr}}, \mathcal{F}_{\mathrm{sep}}, \mathcal{F}_{\mathrm{W}}\right]\right)
\end{equation}
where $[\cdot, \cdot, \cdot]$ denotes the feature concatenation, $\Phi_{\mathrm{fuse}}(\cdot)$ is a two-layer Transformer module, and the input and output of each feature vector maintains the same shape. The visual and the textual features have been augmented through the feature interaction in the Transformer, resulting in the refined features $ \mathcal{F}_{\mathrm{Vr}}^{\prime} \in \mathbb{R}^{|\mathcal{V}| \times d} $ and $ \mathcal{F}_{\mathrm{W}}^{\prime} \in \mathbb{R}^{|\mathcal{W}| \times d} $.

\vspace{5pt}
\noindent \textbf{Adaptive Aggregation.} 
Here, we take out the internal features corresponding to the image global patch and the text global token respectively, and concatenate them together to be transformed as the fusion features $ \mathcal{F}_{\mathrm{U}}\in \mathbb{R}^{d}$ through an MLP module,
we then apply a linear layer to project 
$ \mathcal{F}_{\mathrm{U}} $ into weighting parameters ($w_1, w_2, w_3$) that act as multipliers for $ \mathcal{F}_{\mathrm{Vr}}^{\mathrm{G}}$, 
$ \mathcal{F}_{\mathrm{U}} $ and 
$ \mathcal{F}_{\mathrm{W}}^{\mathrm{G}}$, 
where $ \mathcal{F}_{\mathrm{Vr}}^{\mathrm{G}}$ 
indicates the global BLIP/CLIP visual features, 
$ \mathcal{F}_{\mathrm{W}}^{\mathrm{G}}$ denotes the global BLIP/CLIP textual features. 
The final image-text representation $Q$ is computed as:
\begin{equation}
\small
\label{eq:Q}
    Q = w_1 * \mathcal{F}_{\mathrm{Vr}}^{\mathrm{G}} + w_2 * \mathcal{F}_{\mathrm{U}} + w_3 * \mathcal{F}_{\mathrm{W}}^{\mathrm{G}}
\end{equation}

\subsubsection{The Training Objective}
\label{sec:training_procedure}

For model training, 
we follow previous work and use the batch-based classification (BBC) loss~\cite{vo2019composing}. 
Given a batch size of $B$, the $i$-th query pair ($I_r^i, t^i$) should be close to its positive target $I_t^i$ and far away from the negative instances, which can be formulated as
\begin{equation}
\small
\mathcal{L}=-\frac{1}{B} \sum_{i=1}^B \log \left[\frac{\exp \left[ \kappa\left(Q^i, \mathcal{F}_{\mathrm{Vt}}^i\right) / \tau \right]}{\sum_{j=1}^B \exp \left[ \kappa\left(Q^i, \mathcal{F}_{\mathrm{Vt}}^j\right) / \tau \right]}\right]
\end{equation}
where $\tau=0.01$ refers to the temperature parameter, and $\kappa(\cdot,\cdot)$ denotes the cosine similarity, $Q^i$ is computed by Eq.~\eqref{eq:Q} and $\mathcal{F}_{\mathrm{Vt}}^i = \Phi_{\mathrm{visual}}(I_t^i)$ is the representation of the target image of that query. In practise, to effectively train a model for composed image retrieval, a significant amount of triplet data is often required, unfortunately, collecting and annotating CIR datasets can be time-consuming and costly. In the following section, we describe an automatic pipeline for constructing dataset suitable for CIR training.

\subsection{Dataset Construction}
\label{sec:dataset_construction}

\begin{figure}[t]
    \centering
    \includegraphics[width=.95\textwidth]{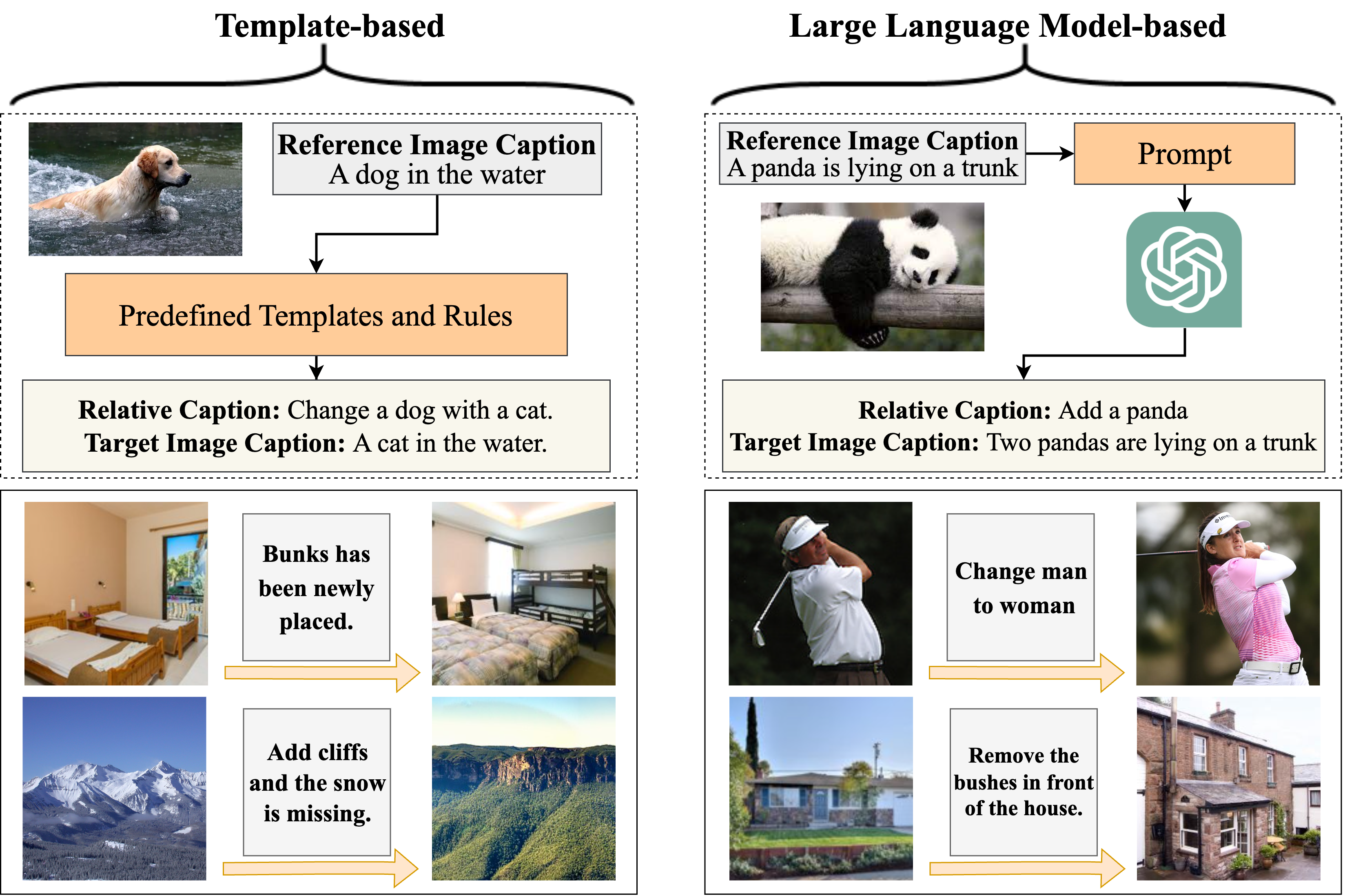}
    \vspace{-5pt}
    \caption{An overview of our proposed dataset construction procedure, 
    based on sentence template~(left), or large language models~(right).}
    \label{fig:data_collection}
    \vspace{-10pt}
\end{figure}

In order to train the CIR model, 
we need to construct a dataset with triplet samples, 
{\em i.e.}, reference image, relative caption, target image.
Specifically, we start from the \textbf{Laion-COCO}\footnote{\href{https://laion.ai/blog/laion-coco/}{https://laion.ai/blog/laion-coco/}} that contains a massive number of image-caption pairs,
and then edit the captions with sentence templates or large-language models~(Sec.~\ref{sec:template}), 
to retrieve the target images~(Sec.~\ref{sec:retrieve}), as shown in Figure~\ref{fig:data_collection}. 
The details are discussed in the following sections.

\subsubsection{Generating Relative Caption}
\label{sec:template}

\noindent {\bf Generation Based on Language Templates. } 
Here, we aim to generate the relative caption based on predefined templates and rules. 
Specifically, we take inspiration from~\cite{liu2021image}, 
and consider eight types of semantic operations, 
namely \textit{cardinality, addition, negation, direct addressing, compare\&change, comparative statement, statement with conjunction and viewpoint}. 
For these operations, it is straightforward to define diverse rules to edit the original caption of Laion-COCO images. Taking the type \textit{compare\&change} as an example, 
we first extract the noun phrases from the captions with a part-of-speech (POS) tagger, 
provided by Spacy~\cite{Honnibal_spaCy_Industrial-strength_Natural_2020}. 
Then, we define the template as: ``replace \{entity A\} with \{entity B\}'', 
where entity A is replaced with other similar noun phrases, measured with the Sentence-Transformers similarity score, {\em i.e.}, we replace the original noun phrase with an alternative noun phrase with similarity ranging from 0.5 to 0.7 measured by all-MiniLM-L6-v2\footnote{\href{https://huggingface.co/sentence-transformers/all-MiniLM-L6-v2}{https://huggingface.co/sentence-transformers/all-MiniLM-L6-v2}}. 
To this end, we acquire the edited image caption, 
which will be later used to retrieve the target image. 
For more implementation details, please refer to our supplementary materials.

\vspace{3pt}
\noindent {\bf Generation Based on Large Language Model. } 
Given the image caption for reference image, we prompt ChatGPT~(gpt-3.5-turbo) to simultaneously generate relative caption and caption of target image, with the {\bf following prompt}:
\textit{I have an image. Carefully generate an informative instruction to edit this image content and generate a description of the edited image. I will put my image content beginning with ``Image Content:''. The instruction you generate should begin with ``Instruction:". The edited description you generate should begin with ``Edited Description:". The Instruction you generate can cover various semantic aspects, including cardinality, addition, negation, direct addressing, compare\&change, comparative, conjunction, spatial relations\&background, viewpoint. The edited description need to be as simple as possible. The instruction does not need to explicitly indicate which type it is. Avoid adding imaginary things. ``Image Content: \{\}''. 
Each time generate one instruction and one edited description only.}

\subsubsection{Target Image Retrieval}
\label{sec:retrieve}

With the target image captions generated by the template-based or LLM-based approach, 
we use a sentence transformer model to extract features from the caption, 
and then we perform a text-only retrieval between the target image caption and the captions of the images in the Laion-COCO pool using cosine similarity. 
The images with their corresponding captions to have similarity scores above the given threshold are kept as candidate target images, resulting in a scalable pipeline for constructing triplet samples, with reference image, relative caption, and target image.

\section{Experiment}
In this section, we first describe the experiment setups
and implementation details~(Sec.~\ref{sec:exp}),
then followed by ablation studies to investigate the applicability of our method and the effectiveness of the core components in our TransAgg model~(Sec.~\ref{sec:abl}),
lastly, we present comparison results to the recent approaches~(Sec.~\ref{sec:comp}). 
{\bf Note that}, there has been several concurrent work on composed image retrieval~\cite{saito2023pic2word, baldrati2023zero, gu2023compodiff, levy2023data}, here, we try to compare with them as fairly as we can, however, there still remain differences on some small experimental details, such as visual and text encoder, embedding dimensions, batch size, {\em etc}.

\subsection{Experimental Setups}
\label{sec:exp}
\noindent \textbf{Training Datasets.}
We construct the training sets by using the data collection pipeline outlined in Section~\ref{sec:dataset_construction},
resulting Laion-CIR-Template and Laion-CIR-LLM,
depending on the adopted approaches. 
Both datasets contain around 16K triplets. We also combined two approaches and construct a 32K dataset, named Laion-CIR-Combined.

\vspace{3pt}
\noindent \textbf{Evaluation Datasets.} 
We evaluate our model on two public benchmarks, namely, CIRR~\cite{liu2021image} and FashionIQ~\cite{wu2021fashion}. \textit{CIRR} comprises approximately 36K triplets that are sampled from generic images obtained from $\rm NLVR^2$~\cite{suhr2018corpus}. 
To mitigate the false negative cases, the author conduct two benchmarks to demonstrate fine-grained retrieval. 
The first one involves a general search using the entire validation corpus as the target search space. 
The second focuses on a subset of six images similar to the query image, based on pre-trained ResNet152~\cite{he2016deep} feature distance.
\textit{FashionIQ} focuses on the fashion domain and is divided into three sub categories, \textit{Dress, Shirt} and \textit{Toptee}. 
It contains more than 30k triplets. 
The reference and target images are matched based on similarities in their titles, 
and each triplet is accompanied by two annotations that are manually generated by human annotators.
{\bf Note that}, in this paper, we consider zero-shot evaluation, 
that is to say, we only train on our automatically constructed training set, 
and directly evaluate on the target benchmarks.

\vspace{3pt}
\noindent \textbf{Evaluation Metrics.} 
We adopt the standard metric in retrieval, {\em i.e.}, $\rm Recall@K$, 
which denotes the percentage of target images being included in the top-$K$ list. 
For CIRR, we also report $\rm Recall_{Subset}@K$ metric, 
which considers only the images within the subset of the query.

\vspace{3pt}
\noindent \textbf{Implementation Details. } 
Our framework is implemented with PyTorch. 
We adopt the same image pre-processing scheme as in CLIP4CIR~\cite{baldrati2022conditioned},
and realize the transformer-based fusion module of 2 layers with 8 heads. 
Regarding the training schedule, AdamW optimizer with a cosine decay is applied. 
The learning rate of the visual and text encoder parameters is initialized to 1e-6, 
while that of the remaining parameters are initialized to 1e-4. 
For visual and text encoders, we use pre-trained BLIP~\cite{li2022blip} w/ViT-B, ViT-B/32 CLIP~\cite{radford2021learning} and ViT-L/14 CLIP~\cite{radford2021learning}. 
The language model used in the process of Laion-CIR-Template dataset construction is all-MiniLM-L6-v2.

\subsection{Ablation Study}
\label{sec:abl}
In this section, we evaluate on FashionIQ and CIRR benchmarks,
to investigate the effectiveness of our proposed dataset construction procedure, compare different pre-trained visual backbones, and ablation studies on the transformer-based fusion, adaptive aggregation. 

\begin{table}[t]
\centering
\scriptsize
\setlength{\tabcolsep}{0.26cm}
\begin{tabular}{c c c c c c c c c}
\hline
 \multicolumn{2}{c}{~}&
 \multicolumn{4}{c}{\textbf{CIRR}}& 
 \multicolumn{3}{c}{\textbf{FashionIQ}} \\
 \cmidrule(lr){3-6} \cmidrule(lr){7-9}
 Backbone&Fine-tuning & R@1& R@5& \text{$\rm R_{Subset}@1$}& Average &R@10& R@50&Average\\
\hline 
  \multirow{3}{5em}{CLIP-B/32} & \ding{56} & 24.46 & 53.61 & 57.81 & 55.71 & 23.91 & 44.68 & 34.30\\
 &only text enc. &27.08 &57.21 &62.70 &59.96 &25.67  &46.43 &35.65\\
 &both&29.30 &60.48 &63.57&62.03&25.15 &46.10 &35.63\\
\hline
 \multirow{3}{5em}{CLIP-L/14} & \ding{56} & 25.04 & 53.98 & 55.33 & 54.66 & 28.57 & 48.29 & 38.43\\ 
 &only text enc. &27.90 &58.27 &60.48 &59.38&30.61 &50.38 &40.50\\
 & both & 33.04 & 64.39 & 63.37 & 63.88 & \underline{{\color{blue}32.63}} & \underline{{\color{blue}53.65}} & \underline{{\color{blue}43.14}}\\
\hline
\multirow{3}{3em}{BLIP}& \ding{56} & 34.89 & 64.75 & 66.34 & 65.55 & 26.95 & 46.10 & 36.53\\ 
& only text enc. & \textbf{{\color{red}38.10}} & \textbf{{\color{red}68.42}} & \textbf{{\color{red}70.34}} & \textbf{{\color{red}69.38}} & 32.07 & 53.26 & 42.67\\
& both & \underline{{\color{blue}37.18}} & \underline{{\color{blue}67.21}} & \underline{{\color{blue}69.34}} & \underline{{\color{blue}68.28}} & \textbf{{\color{red}34.64}} & \textbf{{\color{red}55.72}} & \textbf{{\color{red}45.18}}\\
\hline
\end{tabular}
\vspace{6pt}
\caption{Generalization for different backbones and fine-tuning types on CIRR and FashionIQ. For CIRR, the average column denotes $\rm (Recall@5 + Recall_{Subset}@1) / 2$. For FashionIQ, we report the average $\rm Recall@10$ and 50 of all three categories. Best (resp. second-best) numbers are in red (resp. blue). Refer the reader to supplementary material for more detailed comparison.}
\label{table:ablation_backbone}
\end{table}

\vspace{3pt}
\noindent {\bf Pretrained Backbone and Finetuning. }
We train our TransAgg model on Laion-CIR-Template, and explore various backbones and fine-tuning types. As shown in Table~\ref{table:ablation_backbone}, it can be observed that using BLIP~\cite{li2022blip} model as the visual and text encoder yield the best performance, and fine-tuning more parameters leads better results in most cases. 
In the following experiments, we choose to use BLIP~\cite{li2022blip} model as our visual and text encoder.

\vspace{3pt}
\noindent{\bf Effectness of Individual Modules. } 
We conduct ablation studies on transformer fusion and adaptive aggregation, 
as well as the different ways for constructing dataset, {\em i.e.}, Laion-CIR-Template, 
and Laion-CIR-LLM. As shown in Table~\ref{table:ablation_component_fiq}, 
we can make the following observations:
(i) template-based sentence editing is more effective for dataset construction, 
{\em e.g.}, model C3 vs.~F3; 
(ii) adaptive aggregation has a greater impact than transformer fusion, 
{\em e.g.}, model D1 vs.~D2; 
(iii) finetuning both the text encoder and visual encoder gives better performance, 
similar to the observations in Table~\ref{table:ablation_backbone}, 
{\em e.g.}, model B3 vs.~C3. 
Overall, our results demonstrate positive effects of our module, 
regardless of the fine-tuning type.

\begin{table}[!t]
\scriptsize
\setlength{\tabcolsep}{0.1cm}
\centering
\begin{tabular}{c c c c c|c c|c c|c c|c c}
\toprule
 \multicolumn{5}{c|}{~}&
 \multicolumn{2}{c|}{\bf Shirt}& 
 \multicolumn{2}{c|}{\bf Dress}& 
 \multicolumn{2}{c|}{\bf TopTee}&
 \multicolumn{2}{c}{\bf Average}\\
 Model &  Dataset Const. & Finetune & Fusion & Aggregation & \text{R@10} & \text{R@50} & \text{R@10} & \text{R@50} & \text{R@10} & \text{R@50} & \text{R@10} & \text{R@50}\\ \midrule
 A1& Template & \ding{56} & \ding{52} & \ding{56} & 25.22 & 42.89 & 19.88 & 40.16 & 26.77 & 46.81 & 23.96& 43.29\\ 
 A2& Template & \ding{56} & \ding{56} & \ding{52}&27.43 & 45.24 & 22.21 & 41.40 & 29.07 & 51.66 & 26.24 & 46.10\\
 A3& Template & \ding{56} & \ding{52} & \ding{52}&28.07 & 45.63 & 21.67 & 41.89 & 31.11 & 50.79 & 26.95 & 46.10\\
 \midrule
 B1& Template & text enc. & \ding{52} & \ding{56} & 32.19 & 52.80 & 27.37 & 49.28 & 35.08 & 55.84 &31.55 & 52.64\\ 
 B2& Template & text enc. & \ding{56} & \ding{52}&32.43 & 51.42 & 28.56 & 49.73 & 35.03 & 56.20 & 32.01 & 52.45\\
 B3& Template & text enc. & \ding{52} & \ding{52}&32.83 & 52.31 & 27.67 & 49.38 & 35.70 & 58.08 & 32.07 & 53.26\\
 \midrule
 C1& Template & both & \ding{52} & \ding{56} & 32.78 & 52.55 & 29.65 & 50.22 & 35.90 & 57.27 & 32.78 & 53.35\\ 
 C2& Template & both & \ding{56} & \ding{52}&34.64 & 54.66 & 29.85 & 51.71 & 38.35 & 59.41 & 34.28 & 55.26\\
 C3& Template & both & \ding{52} & \ding{52}&34.84 & 53.93 & 31.28 & 52.75 & 37.79 & 60.48 & 34.64 & 55.72\\
 \midrule
 D1& LLM & \ding{56} & \ding{52} & \ding{56} & 18.74 & 34.45 & 16.41 & 33.57 & 20.50 & 37.43 & 18.55& 35.15\\ 
 D2& LLM & \ding{56} & \ding{56} & \ding{52}&28.21 & 47.60 & 25.88 & 47.05 & 32.99 & 54.56 & 29.03 & 49.74\\
 D3& LLM & \ding{56} & \ding{52} & \ding{52}&31.89 & 48.72 & 25.53 & 46.80 & 32.99 & 54.11 & 30.14 & 49.88\\
 \midrule
 E1& LLM & text enc. & \ding{52} & \ding{56} & 31.55 & 49.76 & 26.23 & 48.29 & 33.86 & 53.70 & 30.55& 50.58\\ 
 E2&  LLM & text enc. & \ding{56} & \ding{52}&32.63 & 52.06 & 28.51 & 49.73 & 35.95 & 57.01 & 32.36 & 52.93\\
 E3& LLM & text enc. & \ding{52} & \ding{52}&32.92 & 52.16 & 28.56 & 49.58 & 36.82 & 58.59 & 32.77 & 53.44\\
 \midrule
 F1& LLM & both & \ding{52} & \ding{56} & 28.85 & 47.99 & 26.62 & 48.14 & 31.06 & 52.01 & 28.84 & 49.38\\ 
 F2& LLM & both & \ding{56} & \ding{52} & 32.04 & 50.74 & 30.39 & 50.87 & 34.93 & 55.79 & 32.45 & 52.47\\ 
 F3& LLM & both & \ding{52} & \ding{52}& 34.64 & 53.58 & 30.84 & 51.22 & 37.99 & 59.15 & 34.49 & 54.65\\ 
 \bottomrule
\end{tabular}
\vspace{6pt}
\caption{Ablation study on FashionIQ. No Fusion means we remove the transformer fusion module, and no Aggregation means we replace adaptive aggregation with a static aggregation utilizing three learnable weight parameters.}
\label{table:ablation_component_fiq}
\end{table}

\subsection{Comparison with State-of-the-art}
\label{sec:comp}
We train our model on the combination of both constructed datasets,
and compare with various zero-shot composed image retrieval methods on CIRR and FashionIQ. 
As shown in Table~\ref{table:compare_sota}, on CIRR dataset, our proposed model achieves state-of-the-art results in all metrics except for Recall@50. 
While on the FashionIQ dataset, our proposed TransAgg model trained on the automatically constructed dataset also falls among the top2 best models, performing competitively with the concurrent work, namely CompoDiff~\cite{gu2023compodiff}.
\textbf{Note that}, CompoDiff has been trained on over 18M triplet samples, 
while ours only need to train on 16k/32k, significantly more efficient than CompoDiff.

\begin{table}[t]
\centering
\scriptsize
\setlength{\tabcolsep}{0.09cm}
\begin{tabular}{c c c c c c c c c c}
\toprule
 \multicolumn{3}{c}{~}&
 \multicolumn{4}{c}{\textbf{CIRR}}& 
 \multicolumn{3}{c}{\textbf{FashionIQ}} \\
 \cmidrule(lr){4-7} \cmidrule(lr){8-10}
 Method&Zero-shot&\# Training triplets & R@1& R@5& R@50&\text{$\rm R_{Subset}@1$}&R@10& R@50&Average\\
\midrule 
Pic2Word~\cite{saito2023pic2word}\fontsize{4pt}{\baselineskip}\selectfont{CVPR’2023} & \ding{52} & - & 23.90 & 51.70 & 87.80 & - & 24.70 & 43.70 & 34.20\\
 PALAVRA~\cite{eccv2022_palavra_cohen}\fontsize{4pt}{\baselineskip}\selectfont{ECCV’2022} &\ding{52}& - &16.62 &43.49&83.95 &41.61 &19.76  &37.25 & 28.51\\
 SEARLE-XL-OTI~\cite{baldrati2023zero}\fontsize{4pt}{\baselineskip}\selectfont{arXiv’2023}&\ding{52}&-&24.87 &52.31&88.58 &53.80&27.61 &47.90 &37.76\\
 CompoDiff w/T5-XL~\cite{gu2023compodiff}\fontsize{4pt}{\baselineskip}\selectfont{arXiv’2023} & \ding{52} & 18m &19.37 & 53.81&90.85 & 28.96 & \textbf{{\color{red}37.36}} & 50.85 & \underline{{\color{blue}44.11}}\\ 
 CASE Pre-LaSCo.Ca.~\cite{levy2023data}\fontsize{4pt}{\baselineskip}\selectfont{arXiv’2023}&\ding{52} &~360k &35.40 &65.78&\textbf{{\color{red}94.63}} &64.29&- &- &-\\
\midrule
\textbf{TransAgg~(Laion-CIR-Template)}& \ding{52} &16k& \textbf{{\color{red}38.10}} & \underline{{\color{blue}68.42}} & 93.51 &\textbf{{\color{red}70.34}} & 32.07 & 53.26 & 42.67\\ 
\textbf{TransAgg~(Laion-CIR-LLM)}& \ding{52} &16k & 36.71 & 67.83 & 93.86& 66.03 & 32.77 & \underline{\color{blue}53.44} & 43.11\\
\textbf{TransAgg~(Laion-CIR-Combined)}& \ding{52} &32k &\underline{{\color{blue}37.87}} & \textbf{{\color{red}68.88}} & \underline{{\color{blue}93.86}}& \underline{{\color{blue}69.79}} & \underline{{\color{blue}34.36}} & \textbf{{\color{red}55.13}} & \textbf{{\color{red}44.75}}\\
\midrule
CLRPLANT w/OSCAR~\cite{liu2021image}\fontsize{4pt}{\baselineskip}\selectfont{ICCV’2021} & \ding{56}&- &19.55 & 52.55&92.38 & 39.20 & 18.87 & 41.53 & 30.20\\
 ARTEMIS~\cite{delmas2022artemis}\fontsize{4pt}{\baselineskip}\selectfont{ICLR’2022}& \ding{56} &-&16.96&46.10&87.73&39.99&26.05&50.29&38.17\\
CLIP4CIR~\cite{baldrati2022conditioned}\fontsize{4pt}{\baselineskip}\selectfont{CVPRW’2022} & \ding{56} &-&38.53 &69.98&95.93&68.19&38.32&61.74&50.03\\
BLIP4CIR+Bi~\cite{liu2023bi}\fontsize{4pt}{\baselineskip}\selectfont{arXiv’2023}& \ding{56} &-&40.15&73.08&96.27&72.10 &43.49 &67.31&55.40\\
CASE~\cite{levy2023data}\fontsize{4pt}{\baselineskip}\selectfont{arXiv’2023}& \ding{56}  &-&48.00 &79.11&97.57&75.88 &48.79&70.68 &59.74\\
\bottomrule
\end{tabular}
\vspace{-6pt}
\caption{Comparasion on CIRR test set and FashionIQ validation set.
 The best and second-best numbers are shown in red and blue respectively. 
 For more detailed comparison, we refer the reader to the supplementary material.}
\label{table:compare_sota}
\vspace{-0.4cm}
\end{table}

\subsection{Failure Cases of Dataset Construction}

\begin{figure}[!htb]
    \centering
    \includegraphics[width=13cm]{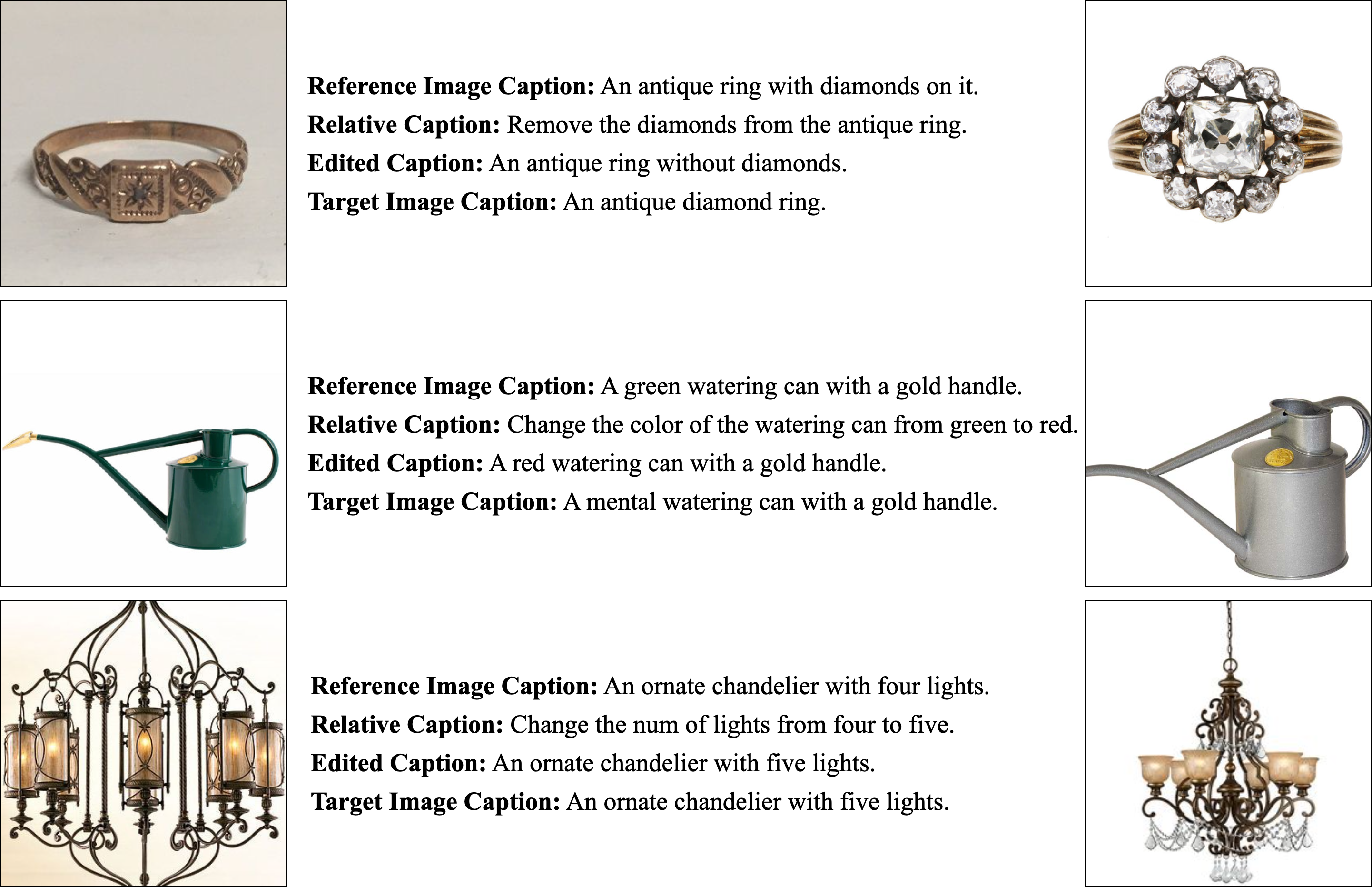}
    \caption{Failure cases of dataset construction. 
    The edited caption and target image caption in the first row have a high similarity score, but their semantic meanings are significantly different. In the second row, we intend to retrieve a red watering can, but a mental watering can is mistakenly retrieved instead. In the third row, the numerical values in both reference image caption and target image caption are incorrect.}
    \label{fig:failure_cases}
    \vspace{-0.2cm}
\end{figure}

There remains limitation on our dataset construction pipeline,
for instance, as shown in the 1st and 2nd row of Figure~\ref{fig:failure_cases}, 
while using sentence transformers for computing sentence similarity,
it may not well capture the crucial information between sentences,
resulting in the failure to retrieve the correct target image. 
Additionally, we use the Laion-COCO as our data corpus, with captions generated automatically, thus can be inaccurate.

\subsection{Qualitative Results for CIR}
\label{sec:qual}
\begin{figure}[!htb]
    \centering
    \includegraphics[width=.95\textwidth]{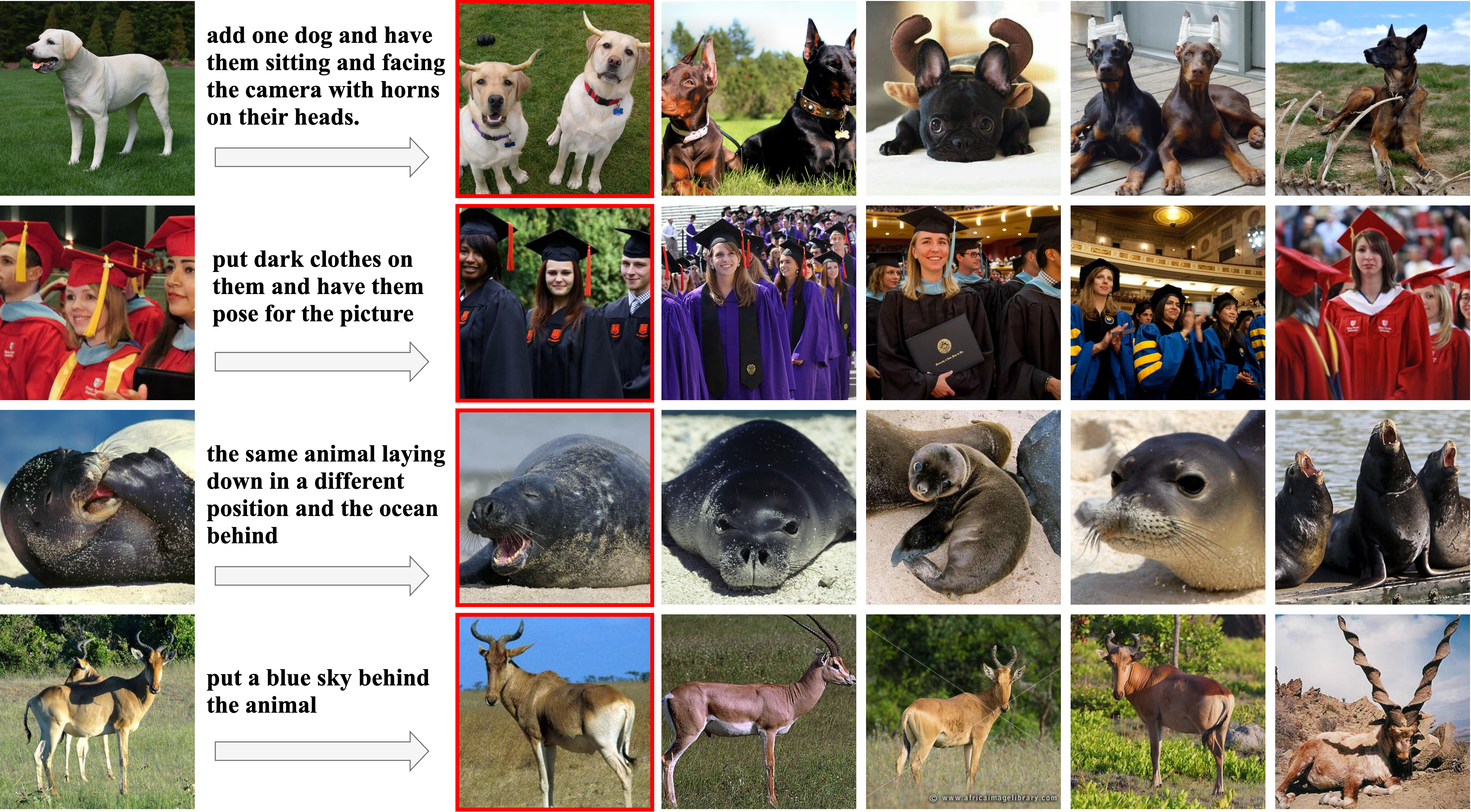}
    \vspace{-5pt}
    \caption{Qualitative results on CIRR.
    From left to right are the reference image, relative caption and the top five retrieved images.
    The ground truth is marked with a red box.
    }
    \label{fig:qualitative_results}
    \vspace{-0.3cm}
\end{figure}

In Figure \ref{fig:qualitative_results}, we show qualitative results on composed image retrieval,
which has only been trained on the automatically constructed dataset, 
without finetuning on the downstream datasets. Each row includes reference image, 
relative caption and the top five retrieved images, 
where the ground truth is marked with a red box. 
The results demonstrate the effectiveness of our proposed method in successfully retrieving the target image. For instance, as shown in the last row, the model must be able to maintain the semantic category of the animal in the reference image, 
and then add a blue sky in order to retrieve the target image.

\section{Conclusion}
In this paper, we propose a retrieval-based pipeline for automatic CIR dataset construction, using the easily-acquired image-caption data on Internet. Specifically, we obtain two different CIR datasets based on templates and large language model. Furthermore, we propose TransAgg, a transformer-based adaptive aggregation model that can effectively integrate information across different modalities. Extensive experiments show that our method performs on par or significant above the existing state-of-the-art (SOTA) models on two public benchmarks and our zero-shot result is sometimes comparable to fully supervised ones.

\noindent \textbf{Acknowledgement.} 
This work is supported by National Key R\&D Program of China (No. 2022ZD0161400). We thank Zechuan Fang and Wenhao Lu for proof-reading.

\clearpage
\bibliography{main}

\begin{thebibliography}{28}
\providecommand{\natexlab}[1]{#1}
\providecommand{\url}[1]{\texttt{#1}}
\expandafter\ifx\csname urlstyle\endcsname\relax
  \providecommand{\doi}[1]{doi: #1}\else
  \providecommand{\doi}{doi: \begingroup \urlstyle{rm}\Url}\fi

\bibitem[Baldrati et~al.(2022)Baldrati, Bertini, Uricchio, and
  Del~Bimbo]{baldrati2022conditioned}
Alberto Baldrati, Marco Bertini, Tiberio Uricchio, and Alberto Del~Bimbo.
\newblock Conditioned and composed image retrieval combining and partially
  fine-tuning clip-based features.
\newblock In \emph{CVPR Workshops}, 2022.

\bibitem[Baldrati et~al.(2023)Baldrati, Agnolucci, Bertini, and
  Del~Bimbo]{baldrati2023zero}
Alberto Baldrati, Lorenzo Agnolucci, Marco Bertini, and Alberto Del~Bimbo.
\newblock Zero-shot composed image retrieval with textual inversion.
\newblock \emph{arXiv preprint arXiv:2303.15247}, 2023.

\bibitem[Brown et~al.(2020)Brown, Mann, Ryder, Subbiah, Kaplan, Dhariwal,
  Neelakantan, Shyam, Sastry, Askell, et~al.]{brown2020language}
Tom Brown, Benjamin Mann, Nick Ryder, Melanie Subbiah, Jared~D Kaplan, Prafulla
  Dhariwal, Arvind Neelakantan, Pranav Shyam, Girish Sastry, Amanda Askell,
  et~al.
\newblock Language models are few-shot learners.
\newblock In \emph{NeurIPS}, 2020.

\bibitem[Bugliarello et~al.(2021)Bugliarello, Cotterell, Okazaki, and
  Elliott]{bugliarello2021multimodal}
Emanuele Bugliarello, Ryan Cotterell, Naoaki Okazaki, and Desmond Elliott.
\newblock Multimodal pretraining unmasked: A meta-analysis and a unified
  framework of vision-and-language berts.
\newblock \emph{Transactions of the Association for Computational Linguistics},
  2021.

\bibitem[Cohen et~al.(2022)Cohen, Gal, Meirom, Chechik, and
  Atzmon]{eccv2022_palavra_cohen}
Niv Cohen, Rinon Gal, Eli~A. Meirom, Gal Chechik, and Yuval Atzmon.
\newblock "this is my unicorn, fluffy": Personalizing frozen vision-language
  representations.
\newblock In \emph{ECCV}, 2022.

\bibitem[Delmas et~al.(2022)Delmas, Rezende, Csurka, and
  Larlus]{delmas2022artemis}
Ginger Delmas, Rafael~S Rezende, Gabriela Csurka, and Diane Larlus.
\newblock Artemis: Attention-based retrieval with text-explicit matching and
  implicit similarity.
\newblock In \emph{ICLR}, 2022.

\bibitem[Dong et~al.(2019)Dong, Li, Xu, Ji, He, Yang, and Wang]{dong2019dual}
Jianfeng Dong, Xirong Li, Chaoxi Xu, Shouling Ji, Yuan He, Gang Yang, and Xun
  Wang.
\newblock Dual encoding for zero-example video retrieval.
\newblock In \emph{CVPR}, 2019.

\bibitem[Goyal et~al.(2017)Goyal, Khot, Summers-Stay, Batra, and
  Parikh]{goyal2017making}
Yash Goyal, Tejas Khot, Douglas Summers-Stay, Dhruv Batra, and Devi Parikh.
\newblock Making the v in vqa matter: Elevating the role of image understanding
  in visual question answering.
\newblock In \emph{CVPR}, 2017.

\bibitem[Gu et~al.(2023)Gu, Chun, Kim, Jun, Kang, and Yun]{gu2023compodiff}
Geonmo Gu, Sanghyuk Chun, Wonjae Kim, HeeJae Jun, Yoohoon Kang, and Sangdoo
  Yun.
\newblock Compodiff: Versatile composed image retrieval with latent diffusion.
\newblock \emph{arXiv preprint arXiv:2303.11916}, 2023.

\bibitem[He et~al.(2016)He, Zhang, Ren, and Sun]{he2016deep}
Kaiming He, Xiangyu Zhang, Shaoqing Ren, and Jian Sun.
\newblock Deep residual learning for image recognition.
\newblock In \emph{CVPR}, 2016.

\bibitem[Hertz et~al.(2022)Hertz, Mokady, Tenenbaum, Aberman, Pritch, and
  Cohen-Or]{hertz2022prompt}
Amir Hertz, Ron Mokady, Jay Tenenbaum, Kfir Aberman, Yael Pritch, and Daniel
  Cohen-Or.
\newblock Prompt-to-prompt image editing with cross attention control.
\newblock \emph{arXiv preprint arXiv:2208.01626}, 2022.

\bibitem[Honnibal et~al.(2020)Honnibal, Montani, Van~Landeghem, and
  Boyd]{Honnibal_spaCy_Industrial-strength_Natural_2020}
Matthew Honnibal, Ines Montani, Sofie Van~Landeghem, and Adriane Boyd.
\newblock {spaCy: Industrial-strength Natural Language Processing in Python}.
\newblock 2020.
\newblock \doi{10.5281/zenodo.1212303}.

\bibitem[Jia et~al.(2021)Jia, Yang, Xia, Chen, Parekh, Pham, Le, Sung, Li, and
  Duerig]{jia2021scaling}
Chao Jia, Yinfei Yang, Ye~Xia, Yi-Ting Chen, Zarana Parekh, Hieu Pham, Quoc Le,
  Yun-Hsuan Sung, Zhen Li, and Tom Duerig.
\newblock Scaling up visual and vision-language representation learning with
  noisy text supervision.
\newblock In \emph{ICML}, 2021.

\bibitem[Klein et~al.(2015)Klein, Lev, Sadeh, and Wolf]{klein2015associating}
Benjamin Klein, Guy Lev, Gil Sadeh, and Lior Wolf.
\newblock Associating neural word embeddings with deep image representations
  using fisher vectors.
\newblock In \emph{CVPR}, 2015.

\bibitem[Levy et~al.(2023)Levy, Ben-Ari, Darshan, and Lischinski]{levy2023data}
Matan Levy, Rami Ben-Ari, Nir Darshan, and Dani Lischinski.
\newblock Data roaming and early fusion for composed image retrieval.
\newblock \emph{arXiv preprint arXiv:2303.09429}, 2023.

\bibitem[Li et~al.(2022)Li, Li, Xiong, and Hoi]{li2022blip}
Junnan Li, Dongxu Li, Caiming Xiong, and Steven Hoi.
\newblock Blip: Bootstrapping language-image pre-training for unified
  vision-language understanding and generation.
\newblock In \emph{ICML}, 2022.

\bibitem[Li et~al.(2020)Li, Yin, Li, Zhang, Hu, Zhang, Wang, Hu, Dong, Wei,
  et~al.]{li2020oscar}
Xiujun Li, Xi~Yin, Chunyuan Li, Pengchuan Zhang, Xiaowei Hu, Lei Zhang, Lijuan
  Wang, Houdong Hu, Li~Dong, Furu Wei, et~al.
\newblock Oscar: Object-semantics aligned pre-training for vision-language
  tasks.
\newblock In \emph{ECCV}, 2020.

\bibitem[Liu et~al.(2021)Liu, Rodriguez-Opazo, Teney, and Gould]{liu2021image}
Zheyuan Liu, Cristian Rodriguez-Opazo, Damien Teney, and Stephen Gould.
\newblock Image retrieval on real-life images with pre-trained
  vision-and-language models.
\newblock In \emph{ICCV}, 2021.

\bibitem[Liu et~al.(2023)Liu, Sun, Hong, Teney, and Gould]{liu2023bi}
Zheyuan Liu, Weixuan Sun, Yicong Hong, Damien Teney, and Stephen Gould.
\newblock Bi-directional training for composed image retrieval via text prompt
  learning.
\newblock \emph{arXiv preprint arXiv:2303.16604}, 2023.

\bibitem[Miech et~al.(2019)Miech, Zhukov, Alayrac, Tapaswi, Laptev, and
  Sivic]{miech2019howto100m}
Antoine Miech, Dimitri Zhukov, Jean-Baptiste Alayrac, Makarand Tapaswi, Ivan
  Laptev, and Josef Sivic.
\newblock Howto100m: Learning a text-video embedding by watching hundred
  million narrated video clips.
\newblock In \emph{ICCV}, 2019.

\bibitem[Ni et~al.(2021)Ni, Huang, Su, Cui, Bharti, Wang, Zhang, and
  Duan]{ni2021m3p}
Minheng Ni, Haoyang Huang, Lin Su, Edward Cui, Taroon Bharti, Lijuan Wang,
  Dongdong Zhang, and Nan Duan.
\newblock M3p: Learning universal representations via multitask multilingual
  multimodal pre-training.
\newblock In \emph{CVPR}, 2021.

\bibitem[Oord et~al.(2018)Oord, Li, and Vinyals]{oord2018representation}
Aaron van~den Oord, Yazhe Li, and Oriol Vinyals.
\newblock Representation learning with contrastive predictive coding.
\newblock \emph{arXiv preprint arXiv:1807.03748}, 2018.

\bibitem[Pan et~al.(2016)Pan, Mei, Yao, Li, and Rui]{pan2016jointly}
Yingwei Pan, Tao Mei, Ting Yao, Houqiang Li, and Yong Rui.
\newblock Jointly modeling embedding and translation to bridge video and
  language.
\newblock In \emph{CVPR}, 2016.

\bibitem[Radford et~al.(2021)Radford, Kim, Hallacy, Ramesh, Goh, Agarwal,
  Sastry, Askell, Mishkin, Clark, et~al.]{radford2021learning}
Alec Radford, Jong~Wook Kim, Chris Hallacy, Aditya Ramesh, Gabriel Goh,
  Sandhini Agarwal, Girish Sastry, Amanda Askell, Pamela Mishkin, Jack Clark,
  et~al.
\newblock Learning transferable visual models from natural language
  supervision.
\newblock In \emph{ICML}, 2021.

\bibitem[Saito et~al.(2023)Saito, Sohn, Zhang, Li, Lee, Saenko, and
  Pfister]{saito2023pic2word}
Kuniaki Saito, Kihyuk Sohn, Xiang Zhang, Chun-Liang Li, Chen-Yu Lee, Kate
  Saenko, and Tomas Pfister.
\newblock Pic2word: Mapping pictures to words for zero-shot composed image
  retrieval.
\newblock In \emph{CVPR}, 2023.

\bibitem[Suhr et~al.(2018)Suhr, Zhou, Zhang, Zhang, Bai, and
  Artzi]{suhr2018corpus}
Alane Suhr, Stephanie Zhou, Ally Zhang, Iris Zhang, Huajun Bai, and Yoav Artzi.
\newblock A corpus for reasoning about natural language grounded in
  photographs.
\newblock \emph{arXiv preprint arXiv:1811.00491}, 2018.

\bibitem[Vo et~al.(2019)Vo, Jiang, Sun, Murphy, Li, Fei-Fei, and
  Hays]{vo2019composing}
Nam Vo, Lu~Jiang, Chen Sun, Kevin Murphy, Li-Jia Li, Li~Fei-Fei, and James
  Hays.
\newblock Composing text and image for image retrieval-an empirical odyssey.
\newblock In \emph{CVPR}, 2019.

\bibitem[Wu et~al.(2021)Wu, Gao, Guo, Al-Halah, Rennie, Grauman, and
  Feris]{wu2021fashion}
Hui Wu, Yupeng Gao, Xiaoxiao Guo, Ziad Al-Halah, Steven Rennie, Kristen
  Grauman, and Rogerio Feris.
\newblock Fashion iq: A new dataset towards retrieving images by natural
  language feedback.
\newblock In \emph{CVPR}, 2021.

\end{thebibliography}
\renewcommand{\appendixname}{Appendix}
\renewcommand{\appendixpagename}{Appendices}
\renewcommand{\appendixtocname}{Appendices}
\renewcommand{\thesection}{\Alph{section}}
\appendix
\section{Appendix}
In this supplementary material, 
we start by detailing the procedure for dataset construction, 
namely, Laion-CIR-Template dataset, 
then we present more detailed experiment comparison.
Additionally, we also show the results for model training on a combined dataset of Laion-CIR-Template and Laion-CIR-LLM. 
Lastly, we present some failure cases from our dataset construction pipeline and several interpretable heatmaps to analyze the reasoning patterns of the TransAgg model.

\subsection{Details on constructing Laion-CIR-Template}
While constructing the Laion-CIR-Template dataset, 
we consider editing the captions from eight semantic aspects,
as detailed in the following sections.

\vspace{3pt}
\noindent \textbf{Cardinality.} 
We identify the reference image captions that contain digits, 
then we construct the relative caption based on the templates shown in Table~\ref{table:templates_for_cardinality}. 
Next, we replace ``num1'' in the reference image caption with ``num2''
or ``a group of'' to get the edited caption. 

\begin{table}[!htb]
\centering
\scriptsize
\begin{tabular}{>{\centering\arraybackslash}m{6cm}}
\hline
 \textbf{Predefined Template} \\
 \hline
 change to \{num2\} \{noun\}.\\
change to a group of \{noun\}.\\
change \{num1\} \{noun\} to \{num2\} \{noun\}.\\
change \{num1\} \{noun\} to a group of \{noun\}.\\
change the num of \{noun\} from \{num1\} to \{num2\}.\\
\hline
\end{tabular}
\vspace{3pt}
\caption{Predefined templates for cardinality type.}
\label{table:templates_for_cardinality}
\end{table}

\noindent \textbf{Addition.} 
We randomly select a noun from the reference image caption, 
and then select another noun that has a similarity score between 0.5 to 0.7 to it. Next, we construct the corresponding relative caption based on the templates listed in Table~\ref{table:templates_for_addition}, 
and obtain the edited caption by adding ``with \{noun\}'' to the reference image caption.

\begin{table}[!htb]
\centering
\scriptsize
\begin{tabular}{>{\centering\arraybackslash}m{6cm}}
\hline
 \textbf{Predefined Template} \\
 \hline
add \{noun\}.\\
\{noun\} has been added.\\
\{noun\} has been newly placed.\\
\hline
\end{tabular}
\vspace{3pt}
\caption{Predefined templates for addition type.}
\label{table:templates_for_addition}
\end{table}

\noindent \textbf{Negation.} 
We randomly select a noun phrase from the reference image caption, 
then use the template defined in Table~\ref{table:templates_for_negation} to construct a relative caption. The edited caption is created by removing the corresponding noun phrase from the reference image caption.

\vspace{3pt}
\noindent \textbf{Direct Addressing.} We randomly select images with a similarity score of 0.5 to 0.7 as target images by comparing their description with the reference images. The caption of the selected target image is referred to as the relative caption.

\begin{table}[!htb]
\scriptsize
\centering
\begin{tabular}{>{\centering\arraybackslash}m{6cm}}
\hline
 \textbf{Predefined Template} \\
 \hline
 no \{noun\_phrase\}.\\
remove \{noun\_phrase\}.\\
\{noun\_phrase\} is gone.\\
\{noun\_phrase\} is missing.\\
\{noun\_phrase\} is no longer there.\\
\hline
\end{tabular}
\vspace{3pt}
\caption{Predefined templates for negation type.}
\label{table:templates_for_negation}
\vspace{-0.4cm}
\end{table}

\vspace{3pt}
\noindent \textbf{Compare \& Change.} 
First, a noun phrase (noun\_phrase1) is randomly selected from the reference image caption. Then, another noun phrase (noun\_phrase2) with a similarity score in the range of 0.5 to 0.7 is chosen as the replacement for noun\_phrase1. The resulting relative caption is generated using the templates defined in Table~\ref{table:templates_for_change}. 
The edited caption is obtained by substituting noun\_phrase1 in the reference image caption with noun\_phrase2.

\begin{table}[h]
\centering
\scriptsize
\begin{tabular}{>{\centering\arraybackslash}m{6cm}}
\hline
 \textbf{Predefined Template} \\
 \hline
 not \{noun\_phrase1\}, but \{noun\_phrase2\}.\\
 replace \{noun\_phrase1\} with \{noun\_phrase2\}.\\
 instead of \{noun\_phrase1\}, show \{noun\_phrase2\}.\\
\hline
\end{tabular}
\vspace{3pt}
\caption{Predefined templates for compare \& change type.}
\label{table:templates_for_change}
\end{table}

\noindent \textbf{Comparative Statement.} 
In this section, we focus on some common adjectives. 
We start by selecting the adjectives from the reference image caption, 
and replacing them with their antonyms to create the edited caption. 
The relative caption is then formed by using the comparative form of the antonym with the noun it modifies.

\vspace{3pt}
\noindent \textbf{Viewpoint.} 
We randomly select a noun from the reference image caption, and use the templates from Table~\ref{table:templates_for_viewpoint} to construct a relative caption. 
We then append either ``small'' or ``big'' to the noun depending on the meaning of the relative caption to create an edited caption.

\begin{table}[h]
\centering
\scriptsize
\begin{tabular}{>{\centering\arraybackslash}m{6cm}}
\hline
 \textbf{Predefined Template} \\
 \hline
 focus on the \{noun\}.\\
 zoom in the \{noun\}.\\
 zoom out the \{noun\}.\\
\hline
\end{tabular}
\vspace{3pt}
\caption{Predefined templates for viewpoint type.}
\label{table:templates_for_viewpoint}
\end{table}

\noindent \textbf{Statement with Conjunction.} This section randomly selects two out of the seven scenarios mentioned earlier and combines them randomly. The final relative caption combines each of their respective relative captions using "and". The edited caption is then modified according to their respective rules.

\subsection{Detailed Experimental Results}
In this section, we present more detailed experimental results.

\subsubsection{Pretrained backbone and finetuning} 
The complete experimental results for different backbone and fine-tuning types on the CIRR and FashionIQ datasets are presented in Table~\ref{table:ablation_cirr_complete} and Table~\ref{table:ablation_fiq_complete}, respectively.

\begin{table}[!htb]
\centering
\scriptsize
\setlength{\tabcolsep}{0.26cm}
\begin{tabular}{c c|c|c|c|c|c|c|c}
\hline
 \multicolumn{2}{c|}{~}&
 \multicolumn{4}{c|}{\text{$\rm Recall@K$}}& 
 \multicolumn{3}{c}{\text{$\rm Recall_{Subset}@K$}} \\
 Backbone&Fine-tuning & K=1& K=5& K=10& K=50& K=1& K=2& K=3\\
\hline
  \multirow{3}{5em}{CLIP-B/32} & \ding{56} & 24.46 & 53.61 & 67.54 & 89.81 & 57.81 & 78.17 & 89.54\\
 &only text enc. &27.08 &57.21 &70.31 &90.39&62.70 &82.41 &92.15\\
 &both&29.30 &60.48 &73.25&92.31&63.57 &82.31 &91.95\\
\hline
 \multirow{3}{5em}{CLIP-L/14} & \ding{56} & 25.04 & 53.98 & 67.59 & 88.94 & 55.33 & 76.82 & 88.94\\ 
 &only text enc. &27.90 &58.27 &71.01 &91.30&60.48 &80.31 &90.75\\
 & both &33.04 &64.39 &76.27 &93.45&63.37 &82.27 &92.22\\
\hline
\multirow{3}{3em}{BLIP}& \ding{56} & 34.89 & 64.75 & 76.24 & 92.22 & 66.34 & 83.76 & 92.92\\ 
& only text enc. & 38.10 & 68.42 & 79.08 & 93.51 & 70.34 & 86.42 & 94.28\\
& both & 37.18 & 67.21 & 77.92 & 93.43 & 69.34 & 85.68 & 93.62\\
\hline
\end{tabular}
\vspace{4pt}
\caption{Generalization for different backbones and fine-tuning types on CIRR.}
\label{table:ablation_cirr_complete}
\end{table}

\begin{table}[!htb]
\begin{scriptsize}
\setlength{\tabcolsep}{0.15cm}
\centering
\begin{tabular}{c c|c c|c c|c c|c c}
\hline
 \multicolumn{2}{c|}{~}&
 \multicolumn{2}{c|}{Shirt}& 
 \multicolumn{2}{c|}{Dress}& 
 \multicolumn{2}{c|}{TopTee}&
 \multicolumn{2}{c}{Average}\\
 Backbone & Fine-tuning & \text{R@10} & \text{R@50} & \text{R@10} & \text{R@50} & \text{R@10} & \text{R@50} & \text{R@10} & \text{R@50}\\
 \hline 
 \multirow{3}{5em}{CLIP-B/32} &  \ding{56} & 25.37 & 42.69 & 19.44 & 42.04 & 26.93 & 49.31 & 23.91& 44.68\\ 
 &only text enc.&27.38 & 45.58 & 21.47 & 43.88 & 28.15 & 49.82 & 25.67 & 46.43\\
 &both&27.48&46.52&20.58&43.28&27.38&48.50&25.15&46.10\\
 \hline
 \multirow{3}{5em}{CLIP-L/14} &  \ding{56} & 29.54 & 47.79 & 23.85 & 44.57 & 32.33 & 52.52 & 28.57& 48.29\\ 
 &only text enc. &30.91 &49.31&27.32 &47.79&33.61 &54.05 &30.61& 50.38\\
 &both &34.79 &53.39&27.71 &49.68&35.39 &57.88 &32.63& 53.65\\
 \hline
 \multirow{3}{3em}{BLIP} &  \ding{56} & 28.07 & 45.63 & 21.67 & 41.89 & 31.11 & 50.79 & 26.95& 46.10\\ 
 &only text enc. &32.83 &52.31&27.67 &49.38&35.70 &58.08 &32.07& 53.26\\
 &both &34.84 &53.93&31.28&52.75&37.79&60.48&34.64& 55.72\\
 \hline
\end{tabular}
\vspace{4pt}
\caption{Generalization for different backbones and fine-tuning types on FashionIQ.}
\label{table:ablation_fiq_complete}
\end{scriptsize}
\end{table}

\subsubsection{Traininig on combination of Laion-CIR-Template and Laion-CIR-LLM} 
In this section, we combine the Laion-CIR-Template dataset with Laion-CIR-LLM dataset to create a new dataset called Laion-CIR-Combined that consists of approximately 32k samples. Subsequently, we train our proposed TransAgg model on the combined dataset and the results are shown in Table~\ref{table:merge_cirr} and Table~\ref{table:merge_fiq}. It can be observed that using more data tends to lead to better results. 

\begin{table}[!htb]
\centering
\scriptsize
\setlength{\tabcolsep}{0.26cm}
\begin{tabular}{c|c|c|c|c|c|c|c}
\hline
 \multicolumn{1}{c|}{~}&
 \multicolumn{4}{c|}{\text{$\rm Recall@K$}}& 
 \multicolumn{3}{c}{\text{$\rm Recall_{Subset}@K$}} \\
 Fine-tuning & K=1& K=5& K=10& K=50& K=1& K=2& K=3\\
\hline
\ding{56} & 35.28 & 64.46 & 76.53 & 92.46 & 65.37 & 83.37 & 92.12\\
 only text enc. &37.87 & 68.88 & 79.60 & 93.86 & 69.79 & 86.09 & 93.93\\
 both&36.71 &67.06 &77.82&93.65&66.25 &84.09 &93.10\\
\hline
\end{tabular}
\vspace{4pt}
\caption{Results on the CIRR test set.}
\label{table:merge_cirr}
\end{table}

\begin{table}[h]
\begin{scriptsize}
\centering
\begin{tabular}{c|c c|c c|c c|c c}
\hline
 \multicolumn{1}{c|}{~}&
 \multicolumn{2}{c|}{Shirt}& 
 \multicolumn{2}{c|}{Dress}& 
 \multicolumn{2}{c|}{TopTee}&
 \multicolumn{2}{c}{Average}\\
 Fine-tuning & \text{R@10} & \text{R@50} & \text{R@10} & \text{R@50} & \text{R@10} & \text{R@50} & \text{R@10} & \text{R@50}\\
 \hline 
 \ding{56} & 30.86 & 49.02 & 24.49 & 45.51 & 32.94 & 54.05 & 29.43 & 49.53\\
 only text enc.&34.45 & 53.97 & 30.24 & 51.91 & 38.40 & 59.51 & 34.36 & 55.13\\
 both&35.03&52.94&32.52&54.34&37.89&59.15&35.15&55.48\\
 \hline
\end{tabular}
\vspace{4pt}
\caption{Results on the FashionIQ validation set.}
\label{table:merge_fiq}
\end{scriptsize}
\end{table}

\subsubsection{Comparison with state-of-the-art} 
Here, we compare our proposed approach with several existing zero-shot composed image retrieval methods on CIRR and FashionIQ datasets,
as shown in Table~\ref{table:sota_cirr_complete} and Table~\ref{table:sota_fiq_complete}.

\begin{table}[!htb]
\centering
\scriptsize
\setlength{\tabcolsep}{0.15cm}
\begin{tabular}{c|c|c|c|c|c|c|c|c|c}
\hline
\multicolumn{1}{c|}{~}&
\multicolumn{1}{c|}{Zero-shot}&
\multicolumn{1}{c|}{\# Training}&
 \multicolumn{4}{c|}{\text{$\rm Recall@K$}}& 
 \multicolumn{3}{c}{\text{$\rm Recall_{Subset}@K$}} \\
 Method& eval & triplets & K=1& K=5& K=10& K=50& K=1& K=2& K=3\\
 \hline
 Pic2Word~\cite{saito2023pic2word}\fontsize{4pt}{\baselineskip}\selectfont{CVPR’2023} &\ding{52}&-& 23.90 & 51.70 & 65.30 & 87.80 & - & - & -\\ 
 PALAVRA~\cite{eccv2022_palavra_cohen}\fontsize{4pt}{\baselineskip}\selectfont{ECCV’2022} &\ding{52}&-& 16.62 & 43.49 & 58.51 & 83.95 & 41.61 & 65.30 & 80.94\\
 SEARLE-XL-OTI~\cite{baldrati2023zero}\fontsize{4pt}{\baselineskip}\selectfont{arXiv’2023} &\ding{52}&-&24.87 &52.31 &66.29 &88.58&53.80 &74.31 &86.94\\
 CompoDiff w/T5-XL~\cite{gu2023compodiff}\fontsize{4pt}{\baselineskip}\selectfont{arXiv’2023}&\ding{52}&18m&19.37 &53.81 &72.02 &90.85&28.96 &49.21 &67.03\\
 CASE Pre-LaSCo.Ca.~\cite{levy2023data}\fontsize{4pt}{\baselineskip}\selectfont{arXiv’2023} &\ding{52}&360k& 35.40 & 65.78 & 78.53 & \textbf{{\color{red}94.63}} & 64.29 & 82.66 & 91.61\\ 
 \hline
 \textbf{TransAgg(Laion-CIR-Template)}&\ding{52} & 16k&\textbf{{\color{red}38.10}}  & \underline{{\color{blue}68.42}}  & \underline{{\color{blue}79.08}}  & 93.51 & \textbf{{\color{red}70.34}}  & \textbf{{\color{red}86.42}}  & \textbf{{\color{red}94.28}} \\ 
 \textbf{TransAgg(Laion-CIR-LLM)}&\ding{52} &16k& 36.71 & 67.83 & 79.03 & 93.86 &66.03& 83.66 & 92.50\\ 
 \textbf{TransAgg(Laion-CIR-Combined)}&\ding{52} &32k& \underline{{\color{blue}37.87}} & \textbf{{\color{red}68.88}} & \textbf{{\color{red}79.60}} & \underline{{\color{blue}93.86}} & \underline{{\color{blue}69.79}} & \underline{{\color{blue}86.09}} & \underline{{\color{blue}93.93}}\\ 
 \hline
 CLRPLANT w/OSCAR~\cite{liu2021image}\fontsize{4pt}
 {\baselineskip}\selectfont{ICCV’2021} &\ding{56}&-& 19.55 & 52.55 & 68.39 & 92.38 & 39.20 & 63.03 & 79.49\\ 
 ARTEMIS~\cite{delmas2022artemis}\fontsize{4pt}{\baselineskip}\selectfont{ICLR’2022} &\ding{56}&- &16.96& 46.10 & 61.31 & 87.73 & 39.99 & 62.20 & 75.67 \\
 CLIP4CIR~\cite{baldrati2022conditioned}\fontsize{4pt}{\baselineskip}\selectfont{CVPRW’2022}&\ding{56}&-&38.53&69.98 &81.86 &95.93 &68.19&85.64 &94.17\\
 BLIP4CIR+Bi~\cite{liu2023bi}\fontsize{4pt}{\baselineskip}\selectfont{arXiv’2023}&\ding{56}&-&40.15&73.08&83.88&96.27&72.10&88.27&95.93\\
 CASE~\cite{levy2023data}\fontsize{4pt}{\baselineskip}\selectfont{arXiv’2023} &\ding{56}&-&48.00 & 79.11 & 87.25 & 97.57 & 75.88 & 90.58 & 96.00\\ 
 \hline
\end{tabular}
\vspace{3pt}
\caption{Comparasion on CIRR test set. The best and second-best numbers are shown in red and blue respectively.}
\label{table:sota_cirr_complete}
\end{table}

\begin{table}[!htb]
\centering
\scriptsize
\setlength{\tabcolsep}{0.06cm}
\begin{tabular}{c|c|c|c c|c c|c c|c c}
\hline
 \multicolumn{1}{c|}{~}&
 \multicolumn{1}{c|}{Zero-shot}&
 \multicolumn{1}{c|}{\# Training}&
 \multicolumn{2}{c|}{Shirt}& 
 \multicolumn{2}{c|}{Dress}& 
 \multicolumn{2}{c|}{TopTee}&
 \multicolumn{2}{c}{Average}\\
 Method& eval & triplets & \text{R@10} & \text{R@50} & \text{R@10} & \text{R@50} & \text{R@10} & \text{R@50} & \text{R@10} & \text{R@50}\\
 \hline
Pic2Word~\cite{saito2023pic2word}\fontsize{4pt}{\baselineskip}\selectfont{CVPR’2023} & \ding{52} &-& 26.20 & 43.60 & 20.00 & 40.20 & 27.90 & 47.40 & 24.70& 43.70\\ 
 PALAVRA~\cite{eccv2022_palavra_cohen}\fontsize{4pt}{\baselineskip}\selectfont{ECCV’2022}& \ding{52} &-&21.49&37.05&17.25&35.94&20.55&38.76&19.76&37.25\\
 SEARLE-XL-OTI~\cite{baldrati2023zero}\fontsize{4pt}{\baselineskip}\selectfont{arXiv’2023} & \ding{52} &-&30.37 &47.49&21.57 &44.47&30.90 &51.76 &27.61& 47.90\\
 CompoDiff w/T5-XL~\cite{gu2023compodiff}\fontsize{4pt}{\baselineskip}\selectfont{arXiv’2023} & \ding{52} &18m&\textbf{{\color{red}38.10}} &\underline{{\color{blue}52.48}}&\textbf{{\color{red}33.91}} &47.85&\textbf{{\color{red}40.07}} &52.22 &\textbf{{\color{red}37.36}}& 50.85\\
 \hline
 \textbf{TransAgg(Laion-CIR-Template)} & \ding{52} &16k&32.83 &52.31&27.67 &49.38&35.70&58.08&32.07&53.26\\
 \textbf{TransAgg(Laion-CIR-LLM)} & \ding{52} & 16k&32.92 &52.16 &28.56 &\underline{{\color{blue}49.58}}&36.82&\underline{{\color{blue}58.59}}&32.77&\underline{{\color{blue}53.44}}\\
 \textbf{TransAgg(Laion-CIR-Combined)} & \ding{52} & 32k&\underline{{\color{blue}34.45}} &\textbf{{\color{red}53.97}}&\underline{{\color{blue}30.24}} &\textbf{{\color{red}51.91}} &\underline{{\color{blue}38.40}} &\textbf{{\color{red}59.51}}&\underline{{\color{blue}34.36}}&\textbf{{\color{red}55.13}}\\
 \hline
 CLRPLANT w/OSCAR~\cite{liu2021image}\fontsize{4pt}{\baselineskip}\selectfont{ICCV’2021} & \ding{56} &-& 17.53 & 38.81 & 17.45 & 40.41 & 21.64 & 45.38 & 18.87& 41.53\\ 
 ARTEMIS~\cite{delmas2022artemis}\fontsize{4pt}{\baselineskip}\selectfont{ICLR’2022}& \ding{56} &-&21.78&43.64&27.16&52.40&29.20&54.83&26.05&50.29\\
CLIP4CIR~\cite{baldrati2022conditioned}\fontsize{4pt}{\baselineskip}\selectfont{CVPRW’2022} & \ding{56} &-&39.99 &60.45&33.81 &59.40&41.41 &65.37 &38.32&61.74\\
 BLIP4CIR+Bi~\cite{liu2023bi}\fontsize{4pt}{\baselineskip}\selectfont{arXiv’2023}& \ding{56} &-&41.76 &64.28&42.09 &67.33&46.61 &70.32&43.49& 67.31\\
 CASE~\cite{levy2023data}\fontsize{4pt}{\baselineskip}\selectfont{arXiv’2023}& \ding{56}  &-&48.48 &70.23&47.44 &69.36&50.18 &72.24 &48.79& 70.68\\
 \hline
\end{tabular}
\vspace{3pt}
\caption{Comparasion on FashionIQ validation set. The best and second-best numbers are shown in red and blue respectively.}
\label{table:sota_fiq_complete}
\end{table}

\subsection{Explainability}
\begin{figure}
    \centering
    \includegraphics[width=13cm]{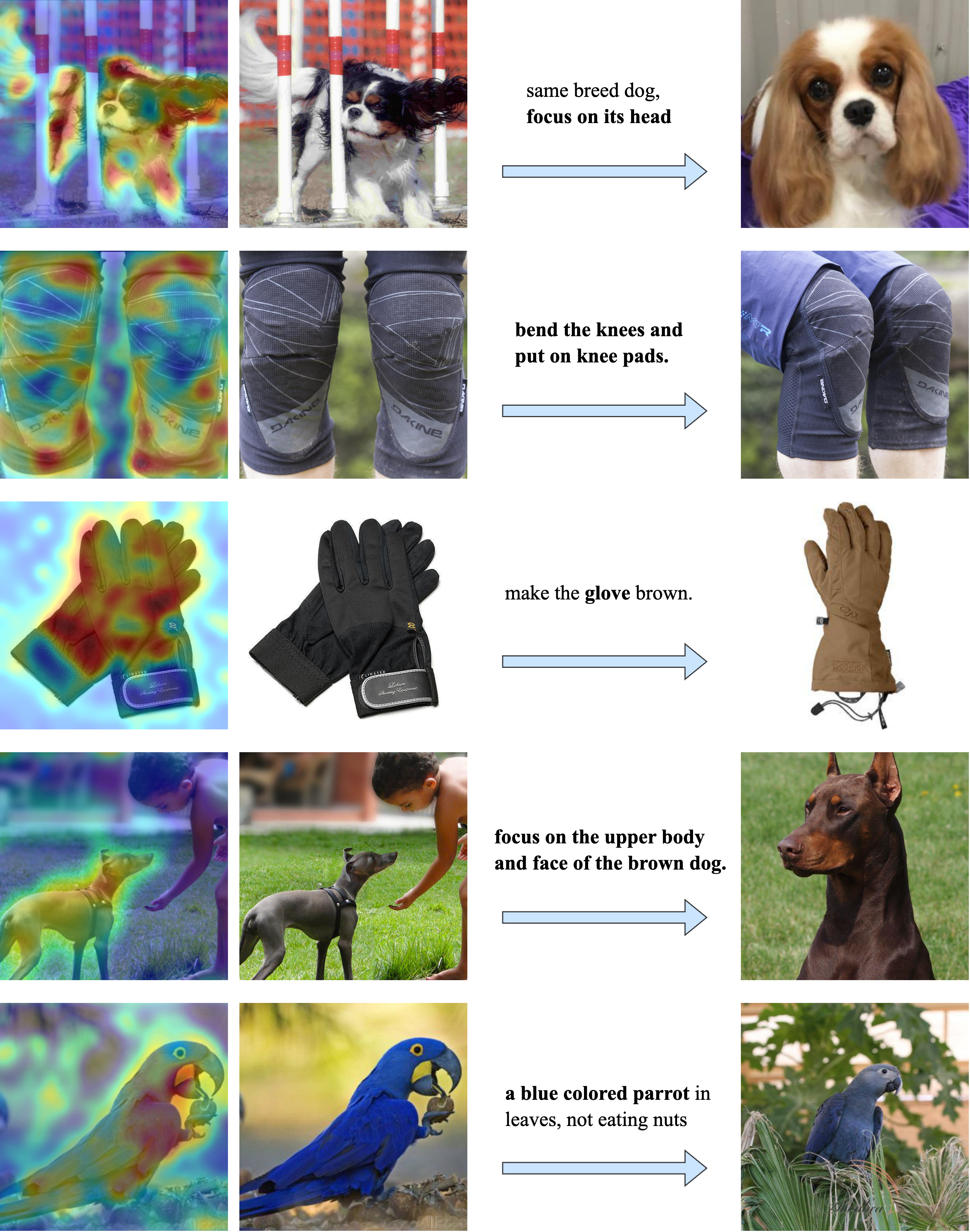}
    \caption{Explainability heatmaps for CIR task. From left to right are the heatmap, reference image, relative caption and the target image. The heatmap is calculated through the attention between the bolded token in the relative caption and other image patches.}
    \label{fiq:heatmap}
\end{figure}

In this section, we present some interpretable examples. As shown in the first row of Figure~\ref{fiq:heatmap}, the relative caption demands a focus on the head of the dog. Correspondingly, the model concentrates most of its attention on the dog. In the second row of Figure~\ref{fiq:heatmap}, the relative caption requires bent knees and knee pads to be worn. Consequently, the model prioritizes the knee and knee pads as the main focal points.

\end{document}